\documentclass[cameraready]{Interspeech}
\usepackage{tabularx}  
\usepackage{booktabs} 
\usepackage{multirow}  
\usepackage{makecell}
\usepackage[T1]{fontenc}
\usepackage[utf8]{inputenc}
\usepackage[table]{xcolor}

\title{Scaling Self-Supervised Speech Models Uncovers Deep Linguistic Relationships: Evidence from the Pacific Cluster}

\author[affiliation={1,2}, orcid=0009-0005-4520-5071]{Minu}{Kim}
\author[affiliation={1}]{Hoirin}{Kim}
\author[affiliation={3}, orcid=0000-0002-3927-6851]{David R.}{Mortensen}

\address{
    $^1$ School of Electrical Engineering, KAIST, Republic of Korea \\
    $^2$ Thomas Lord Department of Computer Science, University of Southern California, USA \\
    $^3$ Language Technologies Institute, Carnegie Mellon University, USA
}

\email{minukim@usc.edu, hoirkim@kaist.ac.kr, dmortens@andrew.cmu.edu}


\usepackage{comment}

\begin{document}

\maketitle

\begin{abstract}
Similarities between language representations derived from Self-Supervised Speech Models (S3Ms) have been observed to primarily reflect geographic proximity or surface typological similarities driven by recent expansion or contact, potentially missing deeper genealogical signals. We investigate how scaling an S3M-based language identification system from 126 to 4,017 languages reshapes this topology, and find a non-linear effect: phylogenetic recovery stays flat up to the 1K scale, but the 4K model undergoes a qualitative shift, resolving both clear lineages and long-term linguistic contact. Most strikingly, a robust Pacific macro-cluster emerges, grouping genealogically unrelated Papuan, Oceanic, and Australian languages—and we trace its driver to a concentrated encoding that captures shared acoustic signatures such as global energy dynamics. These results suggest that massive S3Ms internalize multiple layers of language history, offering a promising perspective for computational phylogenetics and the study of language contact.

\end{abstract}

\section{Introduction}

Our speech reflects not just our lives in the present but a history of linguistic expansion, divergence, and convergence that extends back to human beginnings~\cite{gray2003language,nichols1992linguistic,cavalli1988reconstruction}. 
It seems intuitively likely that representations of our speech, as encoded via self-supervised speech models (S3Ms) also reflect this history and that similarities (and differences) between these representations could provide insight into it. However, investigations have tended to find that S3Ms encode relatively superficial aspects of language relatedness---recent phylogenetic and contact relationships~\cite{toro2025neighbors, 
gutscher2025audio}---and allow researchers to recover limited information regarding the deep history of language groups. In this study, we show that massively scaling up linguistic diversity in the data used to train an S3M for language identification (LID) can yield a qualitative shift in the nature of the representations it produces with respect to linguistic history and deep linguistic relationships. We show that such a model can reliably recover not only close phylogenetic groupings but Sprachbunds~\cite{schaller1997roman} (areas of linguistic convergence) at time depths of several millennia~\cite{gray2009language,bellwood2017first}.

Most existing work has examined models covering hundreds to roughly one thousand languages~\cite{toro2025neighbors,dang2025characterization}.
To investigate the effect of scaling, we compare four LID models, covering 126, 256, 1,024 (1K), and 4,017 (4K) languages respectively, and analyze their embedding spaces across 49 languages spanning diverse families via hierarchical clustering. While phylogenetic recovery plateaus across the 126 to 1K scales, scaling to 4K triggers a pronounced qualitative shift, substantially improving recovery (adjusted rand index: 0.47~$\to$~0.74; normalized mutual information: 0.87~$\to$~0.95), and the 4K model reliably captures several known contact relationships (e.g., Early Chinese cultural sphere, Persian area, and Dravidian area) with high bootstrap confidence.

Among the emergent structures, we identify a particularly striking pattern: Austronesian languages split into two distinct clusters depending on their dispersal history, with the Sundaic and Philippine subgroups forming one cluster, while Oceanic languages group with genealogically unrelated Papuan and Australian languages into a single macro-cluster.
This pattern aligns with the ``Linguistic Melanesia'' convergence~\cite{schapper2020linguistic} and with population genetic evidence for deep interaction across the Pacific~\cite{kayser2006melanesian, lipson2020three, nagele2025impact}, while providing suggestive acoustic evidence for an Australian-Papuan link previously conjectured primarily through archaeology and genomics~\cite{husjahoc2007revealing}.
Focusing on the 1K--4K contrast, we conduct a dimension-level analysis to characterize how this signal is more robustly and distinctively encoded at the 4K scale.

These results suggest that massive scaling in linguistic coverage does not merely scale up the existing representational space, but qualitatively reshapes its geometry, recovering groupings that align with deep phylogenetic and contact relationships otherwise entangled at smaller linguistic scales.


\section{Methods of Analysis}

\subsection{Data}

We select 49 languages spanning diverse families, including Austronesian (Sundaic, Philippine, and Oceanic subgroups), Papuan, Australian, Austroasiatic, Sino-Tibetan, Dravidian, Turkic, Uralic, Indo-European, and Afroasiatic. Audio data are drawn from two publicly available corpora: DoReCo~\cite{paschen2020building}, which provides recordings of 14 primarily low-resource languages (consisting largely of Pacific and Austroasiatic groups), and FLEURS~\cite{conneau2023fleurs}, from which 35 typologically diverse languages are selected.

\subsection{Embedding Extraction}

\begin{figure}[!t]
    \vspace{-5pt}
    \centering
    \includegraphics[width=\columnwidth]{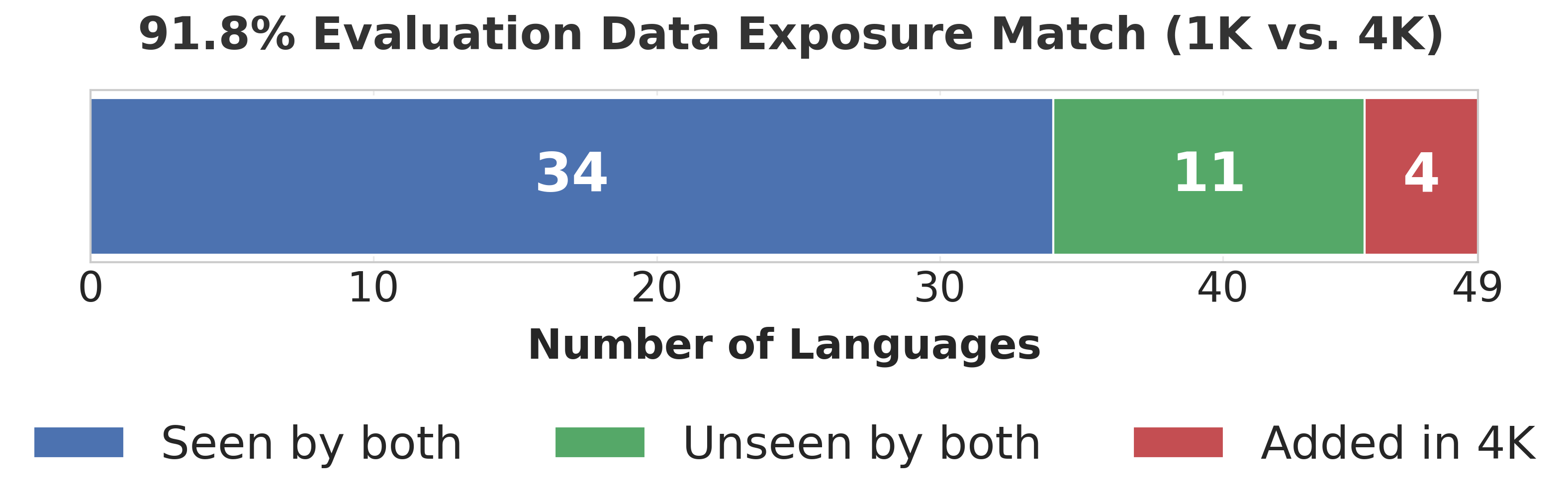}
    \caption{Evaluation data exposure: MMS-LID-1K vs. 4K. Identical seen/unseen status for 91.8\% (45/49) of evaluation languages ensures that performance gaps come from increased linguistic diversity rather than direct exposure.}
    \label{fig:distri}
    \vspace{-5pt}
\end{figure}

We compare four models sharing the same MMS backbone~\cite{pratap2024scaling}: MMS-LID-126\footnote{\url{https://huggingface.co/facebook/mms-lid-126}}, 256\footnote{\url{https://huggingface.co/facebook/mms-lid-256}}, 1,024\footnote{\url{https://huggingface.co/facebook/mms-lid-1024}}, and 4,017\footnote{\url{https://huggingface.co/facebook/mms-lid-4017}}.
Focusing on the 1K--4K contrast to isolate scaling effects from data exposure, we verify that the seen/unseen status is consistent for 45 out of 49 languages (Figure~\ref{fig:distri}). Specifically, 34 are seen and 11 are unseen by both models, with only four languages (Teop, Vera'a, Savosavo, Warlpiri) newly introduced in the 4K model.

For each language $\ell$, we compute a centroid embedding $\mathbf{c}_\ell \in \mathbb{R}^{1280}$ by double-averaging the hidden states of the final transformer layer across all $N_\ell$ constituent audio clips:
\begin{equation}
    \mathbf{c}_\ell = \frac{1}{N_\ell} \sum_{i=1}^{N_\ell} \left( \frac{1}{T_i} \sum_{t=1}^{T_i} \mathbf{h}_{i,t}^{(L)} \right)
\end{equation}
where $\mathbf{h}_{i,t}^{(L)}$ denotes the hidden state at frame $t$ of the $i$-th clip, and $T_i$ is the number of frames in that clip. This process yields a centroid matrix $\mathbf{C} \in \mathbb{R}^{49 \times 1280}$ for each model. To ensure equal weighting during clustering, the dimensions of $\mathbf{C}$ are subsequently standardized to zero mean and unit variance.

\subsection{Hierarchical Clustering and Evaluation}

We apply Ward-linkage agglomerative clustering~\cite{ward1963hierarchical} to the standardized centroids using Euclidean distance, producing a dendrogram over the 49 languages.
Cluster quality is evaluated against established genealogical groupings (defined at the subfamily level) using the adjusted rand index (ARI) and normalized mutual information (NMI)~\cite{hubert1985comparing, vinh2009information}, computed across cluster counts $K \in [2, 20]$.

To assess branch stability of the clades identified in our primary dendrogram, we employ a file-level bootstrap procedure with $B = 1{,}000$ replicates.
In each iteration, we resample the audio files for each language with replacement, recompute centroids, and construct a new dendrogram.
The bootstrap confidence of a clade $S$ is defined as:
\begin{equation}
  \hat{p}(S) = \frac{1}{B} \sum_{b=1}^{B} \mathbf{1}[S \in \mathcal{T}_b]
\end{equation}
where $\mathcal{T}_b$ is the set of clades in the $b$-th bootstrap tree~\cite{felsenstein1985confidence, suzuki2006pvclust}.

\begin{figure}[!t]
    \vspace{-5pt}
    \centering
    \includegraphics[width=0.95\columnwidth]{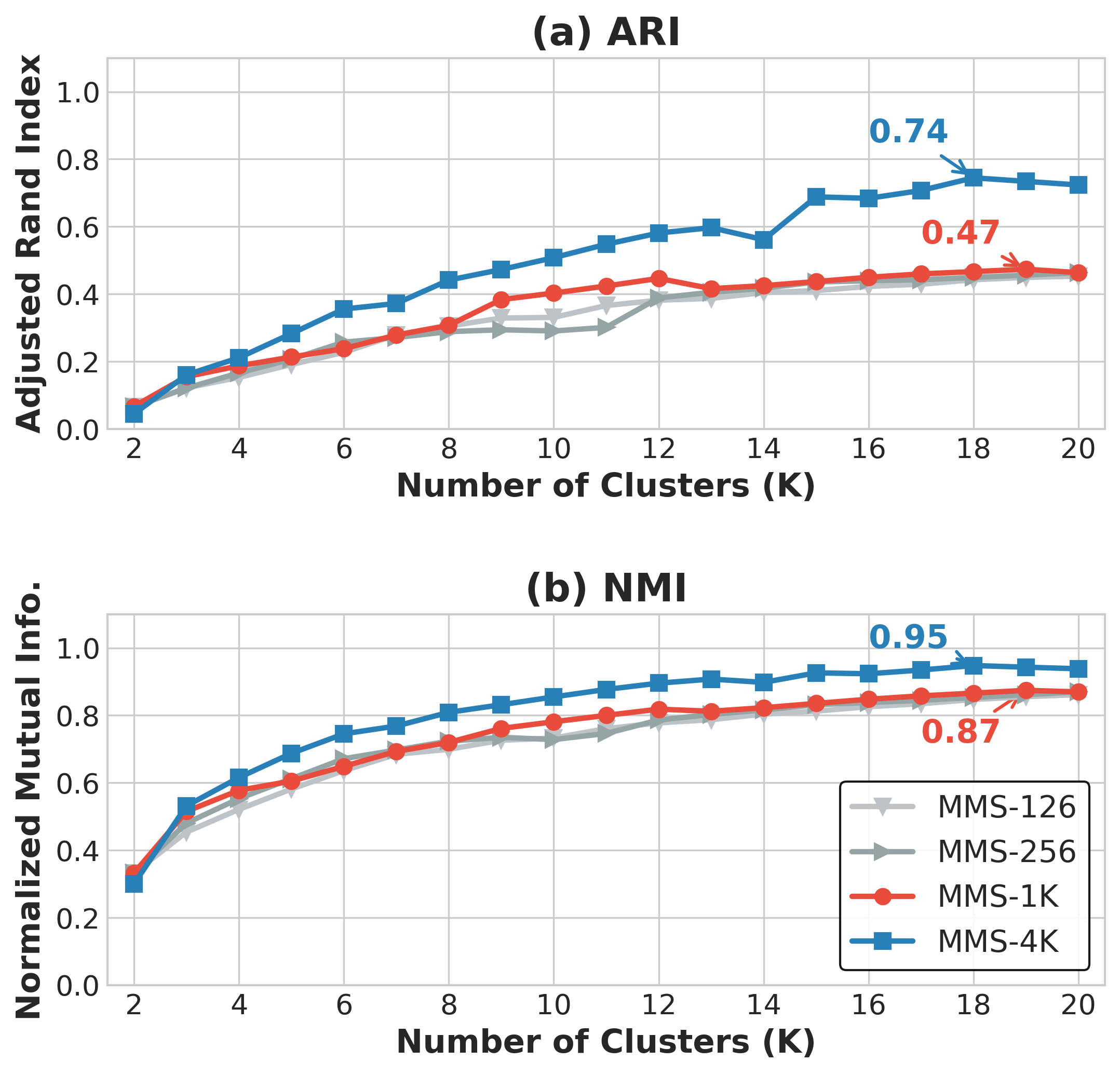}
    \caption{Phylogenetic recovery performance across cluster counts $K \in [2, 20]$ for four model scales. Performance plateaus up to 1K, but shifting to 4K results in a leap that peaks at $K=18$, substantially outperforming all smaller baselines.}
    \label{fig:ari_nmi}
    \vspace{-5pt}
\end{figure}

\subsection{Dimension-Level Analysis}

To identify which latent dimensions drive the observed clustering of Pacific languages (the Pacific--Oceanic--Australian (POA) cluster, as defined in Section~\ref{sec:contact_areal}), we evaluate the discriminative significance of each individual dimension. Specifically, we perform independent $t$-tests comparing POA and non-POA language centroids across all 1,280 dimensions, applying Benjamini--Hochberg false discovery rate (FDR)~\cite{benjamini1995controlling} and Bonferroni~\cite{bonferroni1936teoria} corrections to identify a robust subset of dimensions that carry the most significant discriminative signal.

We then correlate each significant dimension with a set of 30 language-level acoustic features. These features include the dynamic range of energy, as well as the mean and variance of spectral descriptors (e.g., MFCCs, spectral centroid) and temporal dynamics (e.g., zero-crossing rate), extracted directly from the raw audio following well-established standard protocols~\cite{davis1980comparison, tzanetakis2002musical}. To control for varying recording conditions and channel effects, we apply per-utterance root-mean-square (RMS) normalization prior to feature extraction.

For each feature, we compute the \textit{feature frequency}: the proportion of discriminative dimensions with which it significantly correlates. We assess encoding robustness at two stringency levels: (1) an FDR-level analysis, applying FDR-corrected $p < .05$ to both dimension selection and correlation tests, and (2) a conservative Bonferroni-level analysis, applying Bonferroni-corrected $p < .05$ to both stages.

As an independent validation, we conduct Mann--Whitney $U$ tests~\cite{mann1947test} on the same 30 acoustic features between POA and non-POA languages, with effect sizes reported as Cohen's $d$~\cite{cohen2013statistical}, to confirm that the clustering differences in the embedding space correspond to acoustic differences in the raw signal.

\begin{figure}[!t]
    \vspace{-5pt}
    \centering
    \includegraphics[width=\columnwidth]{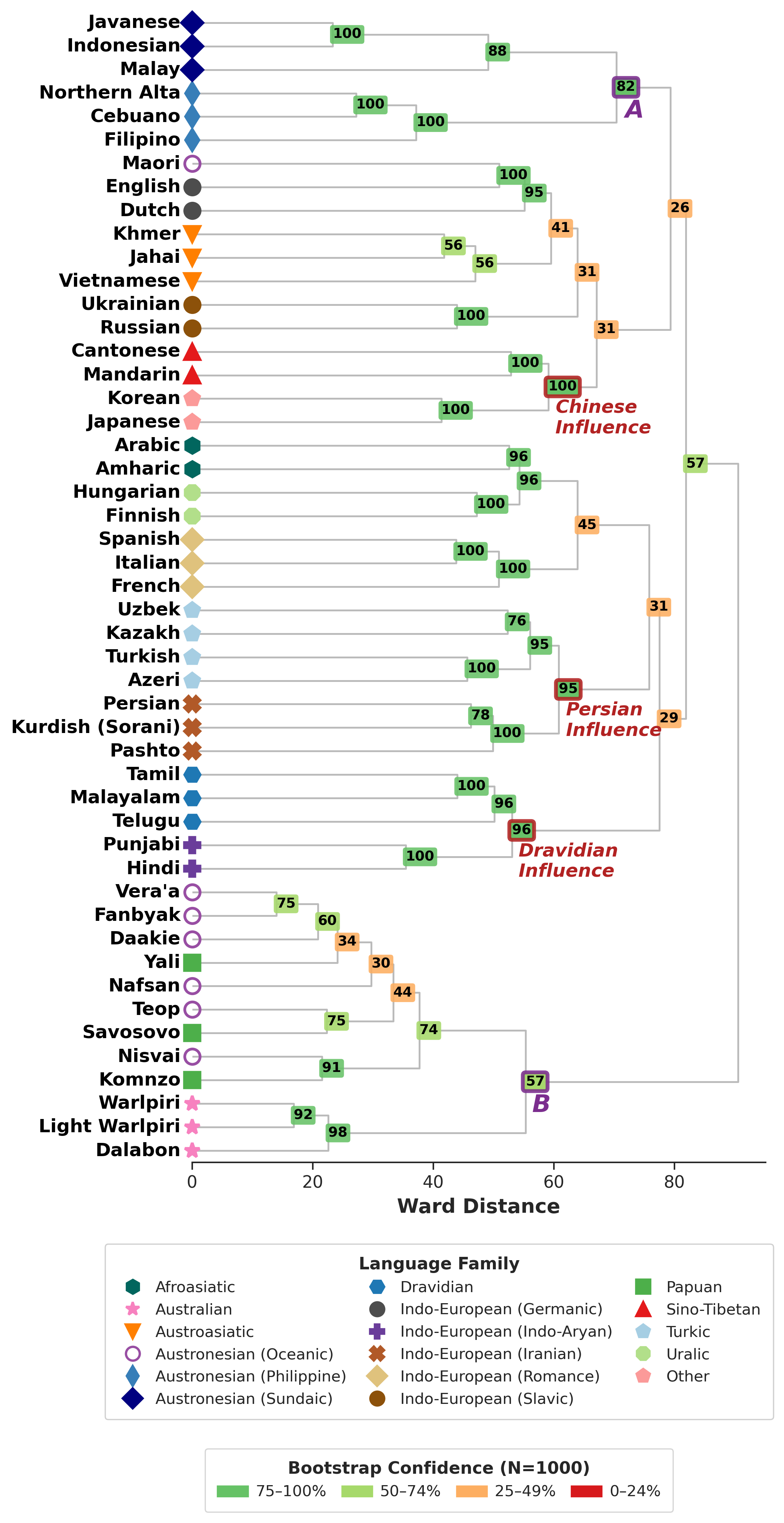}
    \caption{4K model bootstrap consensus dendrogram (1,000 replicates). Branch values indicate support percentages. Red boxes highlight clusters reflecting linguistic contact, including the distinction between Austronesian subgroups A and B. Remarkably, 36 of 37 branches with $>50\%$ confidence align with established phylogenetic or areal groupings.}
    \label{fig:dendro}
    \vspace{-5pt}
\end{figure}

\begin{figure}[!t]
    \vspace{-5pt}
    \centering
    \includegraphics[width=0.95\columnwidth]{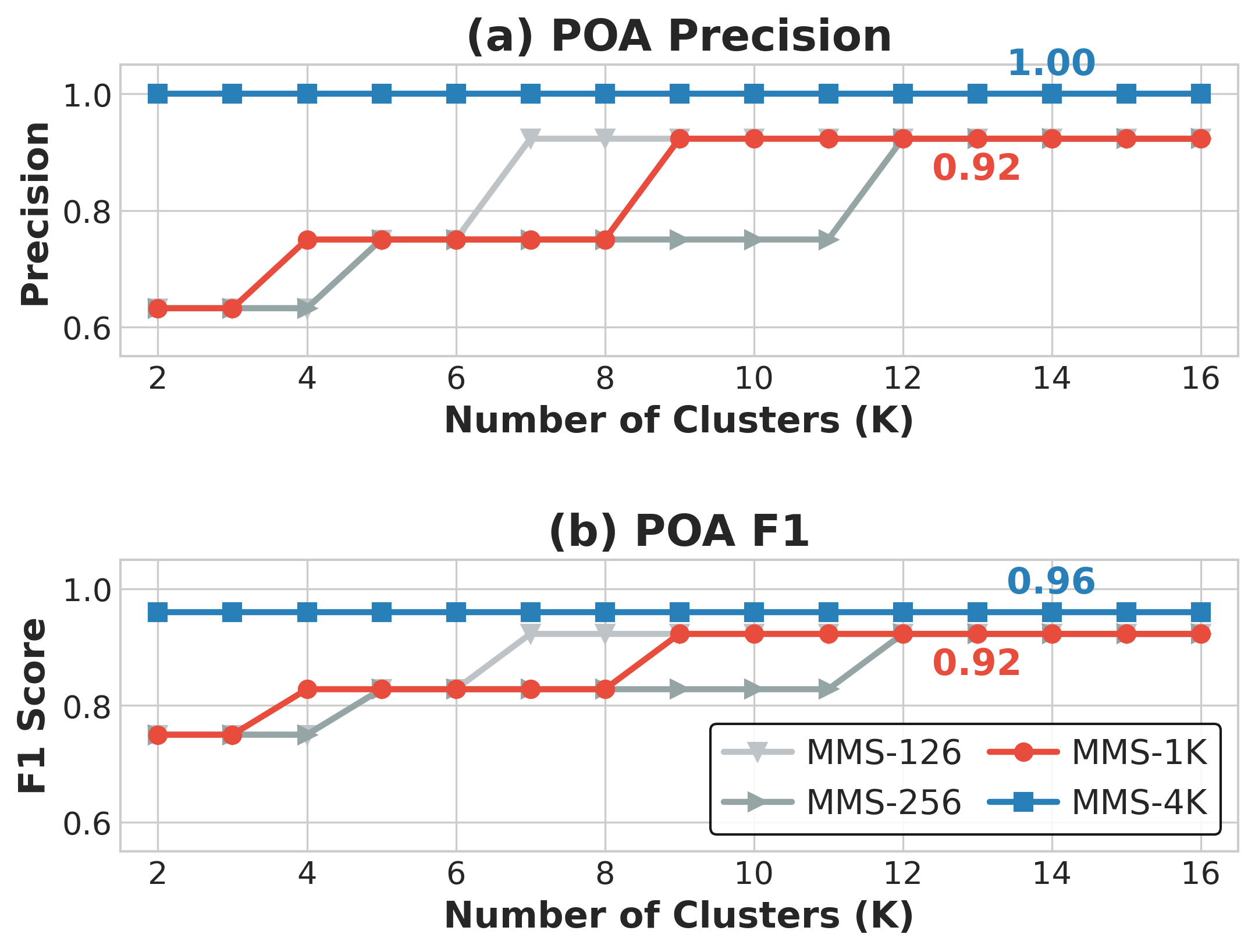}
    \caption{POA cluster quality across $K$. Precision plateaus at 0.92 for 128, 256, and 1K models, but the 4K model achieves perfect Precision (1.00), demonstrating a unique capacity to isolate Pacific languages.}
    \label{fig:poa}
    \label{fig:poa}
    \vspace{-5pt}
\end{figure}

\begin{figure*}[t]
    \vspace{-5pt}
    \centering
    \includegraphics[width=\textwidth]{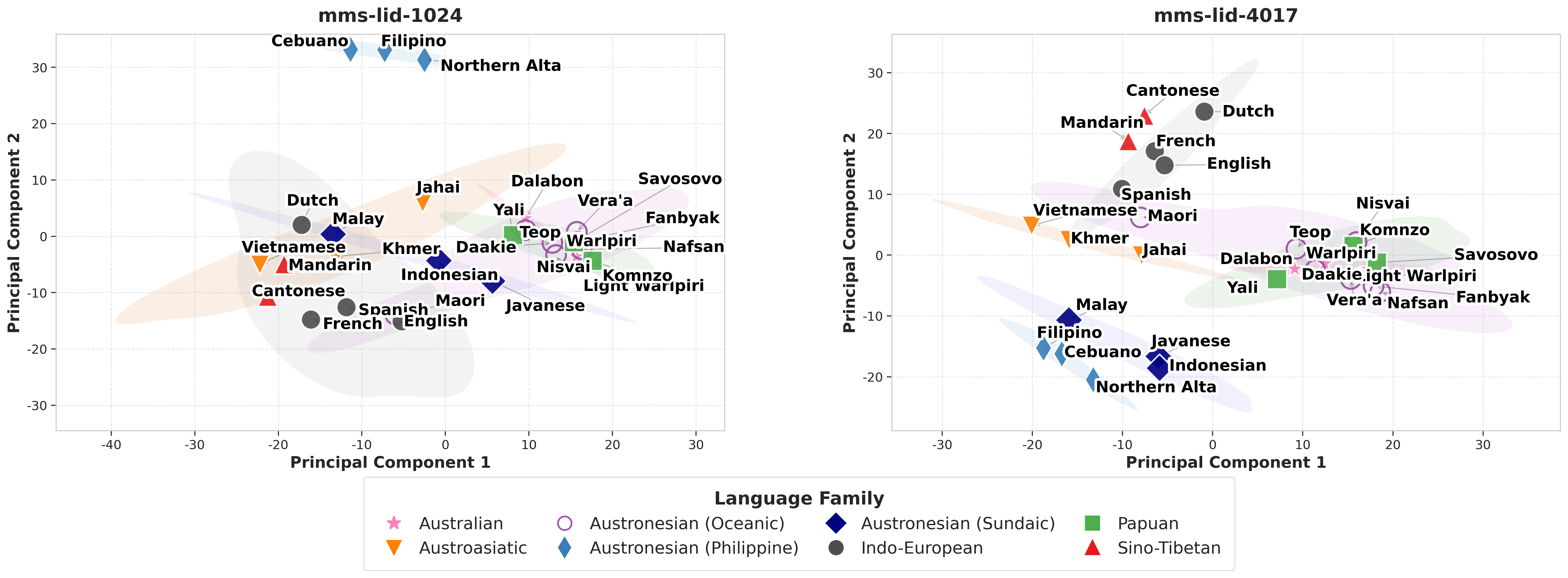}
    \caption{PCA projections of the embedding spaces for 1K (left) and 4K (right) models across selected groups. The 4K model exhibits more clearly delineated family-level boundaries and a more distinct spatial separation of the POA cluster compared to the 1K model.}
    \label{fig:pca}
    \vspace{-5pt}
\end{figure*}

\section{Results}

\subsection{Phylogenetic Recovery}

To evaluate each model's capacity for recovering phylogenetic relationships, we assess its performance at identifying specific genealogical subgroups. Figure~\ref{fig:ari_nmi} illustrates clustering quality via ARI and NMI for $K \in [2, 20]$. 
Notably, performance is stable across the 126 to 1K scales. In contrast, the 4K model displays a performance leap, reaching its peak ARI (0.74) and NMI (0.95) at $K=18$. This aligns with the actual number of subgroups and demonstrates that massive-scale coverage is essential to resolve deep genealogical signals.

\subsection{Linguistic Contact and Areal Patterns}
\label{sec:contact_areal}

Figure~\ref{fig:dendro} presents the 4K consensus dendrogram generated from 1,000 bootstrap replicates.
Nearly all established language families are recovered as coherent subfamily-level clusters, with a sole exception: M\=aori clusters near English (bootstrap: 100\%), likely reflecting colonial contact and bilingualism~\cite{maclagan2004new}; as the sole Polynesian language in our set, it also lies outside the Melanesian sphere of POA.

Beyond genealogy, three documented contact relationships emerge with high bootstrap confidence: (1) a Mandarin--Cantonese--Korean--Japanese cluster (100\%), reflecting Early Chinese influence across East Asia~\cite{sohn2006korean, handel2019sinography}; (2) an Iranian--Turkic group (95\%), consistent with centuries of Persian influence from Anatolia to Xinjiang~\cite{green2019persianate}; and (3) a Dravidian--Indo-Aryan cluster (96\%), aligning with the documented Dravidian substrate in South Asia~\cite{sridhar1981linguistic,sjoberg1992impact}. Overall, 36 of 37 branches (97.3\%) with confidence exceeding 50\% correspond to established phylogenetic or long-term contact groupings (with an Afroasiatic--Uralic cluster as the sole exception).

The dendrogram also reveals two distinct high-level groupings within the Austronesian family, labeled as groups A and B in Figure~\ref{fig:dendro}. Group A (bootstrap: 82\%) comprises Austronesian languages that did not migrate through the New Guinea region, specifically the Philippine and Sundaic subgroups. In contrast, Group B represents a robust macro-cluster (bootstrap: 57\%) where Oceanic Austronesian languages cluster with Papuan and Australian languages, forming the broader Papuan--Oceanic--Australian (POA) group. Within this cluster, Oceanic languages show a particularly strong affinity with Papuan languages, grouping together at 74\% bootstrap confidence.

This structural division reflects the historical trajectory of Austronesian expansion over multiple millennia and typological convergence toward Papuan norms in Melanesia~\cite{schapper2020linguistic, dunn2009contact}. These patterns parallel population genetic evidence of multi-stage migrations and local admixture across the Pacific~\cite{kayser2006melanesian, lipson2020three, nagele2025impact}. While prior research has focused on morphosyntactic borrowing~\cite{ross2001contact}, our results suggest this similarity extends to the acoustic domain. The 4K model's extensive mingling of Oceanic with Papuan and Australian groups indicates internalized shared phonetic patterns, potentially an acoustic signature of regional interaction that remains difficult to resolve through the traditional Comparative Method.

As significantly, the association between Australian languages and Papuan languages provides some of the first acoustic signals consistent with a relationship that has long been conjectured based on archaeological and genetic evidence~\cite{husjahoc2007revealing,malaspinas2016genomic,silcocks2023indigenous}.

\begin{table}[t]
\centering
\small
\caption{Of the total 1,280 dimensions, the first row defines the significant subset used as the denominator for the percentages (\%) of Pearson-correlated dimensions per feature. ($p < .05$)}
\label{tab:integrated_features}
\begin{tabular}{lcccc}
\toprule
 & \multicolumn{2}{c}{MMS-LID-1K} & \multicolumn{2}{c}{MMS-LID-4K} \\
 \cmidrule(lr){2-3} \cmidrule(lr){4-5}
 & FDR & Bonf. & FDR & Bonf. \\
\midrule
\rowcolor[HTML]{F2F2F2} 
\textit{\# Sig. Dimensions} & \textit{257} & \textit{36} & \textit{169} & \textit{25} \\
\midrule
\textit{\textbf{Feature}} & & & & \\ 
Energy dynamic range    & 18.3 & --   & 18.9 & 28.0 \\
MFCC 1 std              & 27.2 & 11.1 &  9.5 & 4.0 \\
MFCC 2 std              & 17.9 &  2.8 & 13.6 & 12.0 \\
MFCC 3 std              & 31.9 & 16.7 & 21.9 & 16.0 \\
MFCC 4 std              & 13.2 & --   &  9.5 & 4.0 \\
Spectral centroid std   & 23.0 & 8.3  & 8.9  & 4.0 \\
Spectral bandwidth std  & 14.4 & --   & 4.7  & 4.0 \\
\bottomrule
\end{tabular}
\vspace{-5pt}
\end{table}

\subsection{Additional Analysis of the POA Cluster}

The 4K model not only clusters POA languages coherently but also isolates them from other families. This emergence of a unified Papuan--Oceanic--Australian space, despite the genealogical diversity of its members, motivates a deeper investigation into the specific acoustic and representational drivers that isolate POA languages from the rest of the world.

\noindent\textbf{Clustering Quality.} We take the POA grouping as given and ask how cleanly each model separates it. Figure~\ref{fig:poa} quantifies this using Precision and F1 scores. The 4K model achieves perfect precision (1.0) across all tested $K$, while the 1K model reaches a maximum of 0.92 at higher cluster counts. F1 scores follow a similar trend: the 4K model holds a stable 0.96 (the minor recall deficit due solely to M\=aori) versus the 1K peak of 0.92.

Figure~\ref{fig:pca} provides complementary visual evidence through PCA projections. In the 1K space, language families exhibit overlap with poorly defined boundaries, and while the POA languages show some initial separation, they remain entangled with other groups. In contrast, the 4K model produces markedly cleaner boundaries between families, with the POA languages forming a tightly delineated and spatially distinct cluster. Notably, this grouping is not driven by corpus origin: non-POA languages from DoReCo (e.g., Jahai, Northern Alta) cluster with their respective families, not co-corpus POA languages.

\noindent\textbf{Dimension Analysis.} Table~\ref{tab:integrated_features} summarizes the distribution of POA-discriminative dimensions. As shown in the first row, the 4K model utilizes a smaller number of significant dimensions compared to the 1K model under FDR (169 vs.\ 257) and Bonferroni (25 vs.\ 36) corrections. Despite this lower count, the 4K model achieves superior POA grouping, suggesting a more concentrated encoding that compresses POA-relevant information into a smaller but more robust set of latent dimensions.

The remainder of Table~\ref{tab:integrated_features}, which reports the most dominant signals among the 30 acoustic features tested, reveals a qualitative shift in these representations. While both models correlate with various spectral descriptors, the 1K model relies more heavily on local spectral fluctuations. In contrast, 4K scaling shifts focus toward more global amplitude dynamics; under the strictest Bonferroni correction, \textit{energy dynamic range} emerges as the most frequent feature in the 4K model (28.0\%), whereas it fails to survive in the 1K model. This transition indicates that linguistic scale pushes the model to integrate broad energy envelopes alongside spectral shape, complementing prior evidence that discrete phoneme-frequency comparisons capture more recent, shallower areal similarity~\cite{kim2025improving}. 

\noindent\textbf{Acoustic Validation.} Table~\ref{tab:acoustic_profile_onecol} independently validates the claim that the S3M representations correspond to genuine acoustic properties. The top features that distinguish POA from non-POA languages in the raw signal, specifically lower spectral variability and higher energy dynamic range, align closely with the features prioritized by the 4K model's discriminative dimensions. This alignment suggests that the 4K model's embedding space reflects phonetic properties of the POA languages.

\begin{table}[t]
\centering
\small
\caption{Significant acoustic differences between POA and Non-POA languages (Mann--Whitney U test). All listed features satisfy $p<.001$ and $|d|>0.5$.}
\label{tab:acoustic_profile_onecol}
\begin{tabular}{lcc}
\toprule
Feature & \makecell{Direction\\(POA)} & \makecell{Effect size\\(Cohen's $d$)} \\
\midrule
Energy dynamic range      & Higher & +0.69 \\
MFCC 1 std                & Lower  & -0.76 \\
MFCC 2 std                & Lower  & -0.88 \\
MFCC 3 std                & Lower  & -1.17 \\
MFCC 4 std                & Lower  & -0.76 \\
Spectral centroid std     & Lower  & -0.61 \\
Spectral bandwidth std    & Lower  & -0.61 \\
\bottomrule
\end{tabular}
\vspace{-10pt}
\end{table}

\section{Conclusion}

This study demonstrates that increasing the linguistic diversity of S3M training data enables models to recover deep linguistic structures, including genealogical subgroups and historical contact patterns. Beyond validating established relationships, the emergence of the Papuan--Oceanic--Australian cluster points toward long-term convergence in Melanesia, providing an acoustic perspective on regional interactions. These findings offer a powerful computational approach for uncovering latent linguistic interactions difficult to resolve via traditional methods.

\section{Acknowledgments}
This work was supported by Institute of Information \& communications Technology Planning \& Evaluation (IITP) grant funded by the Korea government(MSIT) (No.RS-2025-02215393).

\section{Generative AI Disclosure}
Generative AI tools were used to improve the clarity and grammar of the manuscript and to assist with portions of the code. All outputs were reviewed and verified by the authors, who take full responsibility for the content.

\bibliographystyle{IEEEtran}
\bibliography{mybib}

\end{document}